\documentclass[journal,twoside,web]{ieeecolor}
\usepackage{url}
\usepackage{hyperref}
\usepackage{doi}
\usepackage{generic}
\usepackage{cite}
\usepackage{amsmath,amssymb,amsfonts}
\usepackage{algorithmic}
\usepackage{graphicx}
\usepackage{algorithm,algorithmic}
\hypersetup{hidelinks=true}
\usepackage{textcomp}
\def\BibTeX{{\rm B\kern-.05em{\sc i\kern-.025em b}\kern-.08em
    T\kern-.1667em\lower.7ex\hbox{E}\kern-.125emX}}

\usepackage{amsmath}
\usepackage{amssymb}
\usepackage{caption}
\usepackage{subcaption} 
\usepackage{textcomp}
\usepackage{multirow}
\usepackage{booktabs}
\usepackage{makecell}
\usepackage{flafter}
\usepackage{arydshln}

\begin{document}
\title{TLNM: Externally Validated Tooth Detection, Numbering and Segmentation from Smartphone Photographs Using Mask R-CNN}
\author{Arash Nedaei \href{https://orcid.org/0009-0002-8594-8986}{\textsuperscript{\includegraphics[scale=0.06]{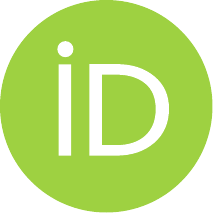}}} , 
    Henna Tiensuu \href{https://orcid.org/0000-0002-6026-9509}{\textsuperscript{\includegraphics[scale=0.06]{orcid.pdf}}},
    Elina Väyrynen \href{https://orcid.org/0009-0000-6890-3810}{\textsuperscript{\includegraphics[scale=0.06]{orcid.pdf}}},
    Saujanya Karki \href{https://orcid.org/0000-0002-3160-8137}{\textsuperscript{\includegraphics[scale=0.06]{orcid.pdf}}},
    and Jaakko Suutala \href{https://orcid.org/0000-0002-6605-0057}{\textsuperscript{\includegraphics[scale=0.06]{orcid.pdf}}}
\thanks{\textbf{Preprint Notice:} This work has been submitted to the IEEE for possible
publication. Copyright may be transferred without notice, after which
this version may no longer be accessible.\\}
\thanks{Arash Nedaei, Henna Tiensuu, and Jaakko Suutala are with Biomimetics and Intelligent Systems Group, Faculty of Information Technology and Electrical Engineering, University of Oulu, Oulu, Finland (e-mail: arash.nedaeijanbesaraei@oulu.fi)}
\thanks{Elina Väyrynen and Saujanya Karki are with the Research Unit of Population Health, Faculty of Medicine, University of Oulu, Oulu, Finland}
}

\maketitle

\begin{abstract}
Oral health issues affect billions of people globally, but the cost and limited access to professional dental care hinder preventive oral healthcare. Research relies on clinical-grade sensors, unavailable for public self-screening. This study introduces a tooth localisation and numbering model for smartphone photographs. We developed a customised Mask Region-based Convolutional Neural Network pipeline trained on 1,272 annotated smartphone images. To address variability in patient-generated health data, the pipeline incorporates two domain-informed mechanisms: a masked gray-world white-balancing algorithm to mitigate artificial colour casts and an anatomically constrained detection layer to enforce structural validity and suppress false positives. The system was evaluated using internal and external testing, descriptive ablation study and training stability analysis. On the internal test set, the model achieved an instance-mask $\text{AP}_{50}$ of 0.818, class-aware PQ of 0.780, and operational F1 of 0.884. On the external dataset, the model achieved an instance-mask $\text{AP}_{50}$ of 0.901, class-aware PQ of 0.832, and operational F1 of 0.928 despite differences in population, sensors, and acquisition protocols. The inference pipeline is available as an open-source, containerised API. These results demonstrate that consumer-grade smartphone imagery can support automated tooth-level anatomical mapping, offering a scalable, potentially low-cost foundation for remote screening and tele-dentistry in resource-constrained environments.
\end{abstract}


\section{Introduction}
\label{sec:introduction}
\IEEEPARstart{G}{lobally}, approximately 3.5 billion people are affected by oral health problems, with untreated cavities being the most prevalent health issue worldwide. Although early detection is vital for successful treatment, obstacles such as high expenses, remote locations, and restricted access to dental care professionals impede timely intervention, particularly in areas with limited resources \cite{who2022GlobalOralHealth}. The widespread use of smartphones, along with advances in deep learning (DL), offers a valuable opportunity to make dental screening more accessible through teledentistry and mobile health (mHealth) technologies. 

However, the transition from clinical dental diagnostics to smartphone-based analysis presents significant engineering challenges. Existing research on automated tooth detection predominantly relies on high-quality clinical imaging modalities such as panoramic radiographs (OPGs), bitewing radiographs, and intraoral photographs captured using specialised cameras and lip retractors. Although these studies demonstrate high accuracy, they fail to account for noise, variable lighting, motion blur, and soft tissue occlusions inherent in patient-generated health data (PGHD) acquired via smartphones. Furthermore, models in this domain are predominantly developed and assessed using data from a single source, raising concerns regarding their generalisability to diverse patient populations.

This study addresses these gaps by introducing tooth localisation and numbering model (TLNM), a DL pipeline specifically optimised for unconstrained smartphone photography. To overcome the high variability of dental PGHD, we moved beyond the standard "black-box" implementations by embedding domain-specific anatomical heuristics directly into the architecture.

This study primarily contributes (1) a domain-informed masked gray-world white-balancing preprocessing algorithm to normalise illumination-related colour variation, (2) an anatomically constrained detection layer that suppresses implausible tooth predictions, and (3) internal stability analysis and independent external validation. Moreover, the TLNM inference pipeline is provided as an open-source, containerised API to support future research and address reproducibility challenges in medical artificial intelligence.

A preliminary, clinically oriented analysis of this project was previously reported as a Research Square preprint \cite{vayrynen2025EvaluationDeepLearning}, and a corresponding manuscript has been submitted for publication. The previous report focused primarily on the feasibility and performance of an earlier version of the Mask R-CNN tooth detection and numbering pipeline. The present study extends that work by adding segmentation focus, introducing a domain-informed masked gray-world white-balancing procedure and an improved anatomically constrained detection layer embedded in the model. In addition, the present study employs a comprehensive evaluation framework, and releases the inference pipeline as an open-source containerised API.

\section{Related Work}
\label{sec:related_work}

The integration of DL into dental diagnostics has rapidly progressed. Huang et al. \cite{huang2023ReviewDeepLearningb} highlighted in their review that DL algorithms have demonstrated remarkable success in dental image segmentation, recognition, and detection of oral conditions and abnormalities, which enables more accurate identification of early-stage dental diseases. This rapid advancement is supported by a growing body of research focused on applying AI, to improve oral disease detection, treatment planning, and workflow efficiency in dental practices \cite{chong2025IntegratingArtificialIntelligencea}. 

\subsection{Deep Learning in Dental Imaging}
Dental imaging has extensively used deep learning, employing well-known architectures like convolutional neural networks (CNNs), Faster R-CNN, YOLO, and U-Net \cite{sivari2023DeepLearningDiagnosisa}. A comprehensive mapping review by Sohrabniya et al. \cite{sohrabniya2025ExploringDecadeDeep}, covering more than 1,000 publications up to September 2023, reported that 84.4\% of the reviewed studies used image-based data. However, the literature has strongly concentrated on tasks such as caries detection, oral cancer diagnosis, and radiological image segmentation, whereas tooth detection and classification have received comparatively less attention.

Within tooth-level analysis, instance segmentation frameworks such as Mask R-CNN have demonstrated strong capability for delineating individual tooth boundaries in radiographic images \cite{wang2024CoMaskRCNNCollaborative}, providing a methodological basis for anatomical localisation and segmentation. Nevertheless, radiographic and other clinical imaging modalities remain dominant; fewer than 4\% of the studies reviewed by Sohrabniya et al. used two-dimensional extraoral images \cite{sohrabniya2025ExploringDecadeDeep}.

Extending these methods to smartphone-captured oral images introduces additional challenges, including variable sensors, variable illumination, motion blur, inconsistent viewing angles, and soft tissue occlusion inherent in patient-generated photography.

\subsection{Automated Tooth Localisation and Numbering}
Precise tooth localisation and numbering are prerequisites for an automated diagnostic pipeline, as they provide anatomical indexing for downstream diagnostic tasks. Historically, dentists performed this manually, making it time-consuming and error-prone.

Recent efforts have sought to automate this process using CNNs, although the literature is heavily skewed toward radiographic images. A systematic review by Maganur et al. \cite{maganur2024DevelopmentArtificialIntelligence} reported that CNN-based models could achieve high precision ($>98\%$) for tooth detection and numbering on dental radiographs. This data modality offers high contrast and structural clarity, which are not present in standard RGB images.

As in radiographic analysis, accurate tooth localisation remains essential for automated diagnostics in optical imaging. For example, Chen et al. \cite{chen2024DeepPlaqDentalPlaque} developed "DeepPlaq" to automate plaque indexing. They used YOLOv8 for localisation, achieving a detection mean Average Precision (mAP) of 0.941 and enabling subsequent tooth numbering using spatial coordinate algorithms. However, the study relied on high-quality Nikon D90 DSLR images which were captured under controlled conditions with cheek retractors and standardised views.

In optical dental imaging, studies distinguish pathology detection from anatomical tooth indexing. For exmaple, Adnan et al. \cite{adnan2024DevelopingAIbasedApplication} developed a YOLOv5s-based smartphone application for localising caries-affected teeth, achieving 90.7\% precision, but its bounding boxes targeted teeth with visible decay rather than systematically identifying all teeth. In contrast, Yoon et al. \cite{yoon2024AIbasedDentalCaries} used a cascade R-CNN on full intraoral photographs to localise and recognise tooth numbers, achieving an overall mAP of 0.880. Although this demonstrates feasibility, the dataset was acquired by dentists using DSLR cameras under controlled procedures. Consequently, automated anatomical indexing from unconstrained smartphone-acquired oral photographs remains comparatively underexplored.

\subsection{Methodological Gaps and External Validation}
A synthesis of the current literature reveals two significant methodological shortcomings that hinder the practical implementation of automated dental indexing in smartphone-captured images. First, existing models are predominantly tailored for professional hardware and strictly controlled clinical settings, whereas the transition to PGHD introduces challenges such as unpredictable illumination, motion blur, overlapping tooth boundaries and clinical occlusions. Second, related studies often lack external validation, raising concerns regarding the generalisability of models across patient populations, data collection instruments, and acquisition procedures. Together, these limitations leave the robustness of automated dental indexing on heterogeneous PGHD images insufficiently established.

This study directly addresses these limitations by presenting a domain-aware DL pipeline for tooth localisation, segmentation, and numbering, specifically optimised for the noise and variability of smartphone-captured images. In addition, the pipeline underwent comprehensive validation through training stability assessment and testing on an independent external dataset. This evaluation protocol provides evidence of its stability and applicability beyond the development data.

\section{Materials and Methods}
\label{sec:methods}

This study was conducted and reported according to the CLAIM guidelines \cite{tejani2024ChecklistArtificialIntelligence} to ensure reproducibility of the results. Data management adhered to the permissions granted by the Finnish Social and Health Data Permit Authority (Findata, THL/6268/14.02.00/2021).

\subsection{Datasets}
\subsubsection{Internal Dataset}
\begin{figure}[t]
    \centering
    \includegraphics[width=\linewidth]{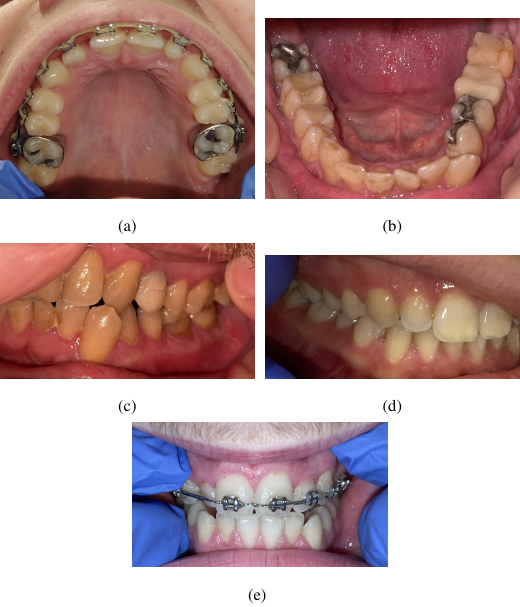}
    \caption{Sample images from the internal test split of the DigiLeap dataset. (a) Upper view, (b) Lower view, (c) Front-left view, (d) Front-right view, (e) Front view.}
    \label{fig:digileap-internal-testset-samples}
\end{figure}

This study utilized data that was collected in the DigiLeap of Oral Health project \cite{UniOulu2021DigileapOralHealth}, an ERDF-funded initiative (EURA 2014/11292/09 02 01 01/2021/PPL). The DigiLeap dataset includes dental health questionnaire, oral cavity images, and clinical examination data. Data were collected in 2022 from 448 participants aged 13–78 years. The image acquisition protocol reflected natural smartphone use without specialised tools or controlled conditions, with minimal guidelines for capturing photographs. Participants captured RGB oral cavity photographs using their own or the surveyor's smartphone and uploaded them via a web-based application. Each participant was expected to provide five images: two occlusal (upper and lower jaws), two lateral (premolar and molar regions), and one frontal (incisor–canine region) view. However, owing to missing and non-applicable images, the DigiLeap dataset contained images from 326 participants, including 229 secondary school students from northern Finland and 97 Finnish adults who were surveyed at the University of Oulu and its partner organisations. Figure \ref{fig:digileap-internal-testset-samples} shows samples from the five views in the dataset.

In the DigiLeap project, annotations were performed in Label Studio \cite{LabelStudio} by four dentists and two senior dental students trained and calibrated using 30 reference images. Each visible tooth was delineated with a polygon and assigned a number based on the World Dental Federation (FDI) notation. In the DigiLeap project, the annotations were exported in JSON format for further analysis.

Parameter-level missingness analysis revealed that the records of tooth numbers and polygon coordinates in the dataset were complete. However, moving to the participant level, only 44.17\% of the participants provided a complete set of images. Table \ref{tab:participant_level_completeness} presents the detailed distribution of image counts per participant across the dataset. No mitigation of missing views was attempted to avoid introducing bias.

In the next step, a stratified random sample of 110 images (~10\% of the dataset) was analysed to estimate the missingness of tooth polygon annotations. Missingness probability was calculated as

\begin{equation} 
    \label{eq:missingness_rate}     
    \hat{p}_{missing} = \frac{\sum_{i=1}^n{m_i}}{\sum_{i=1}^n(m_i + a_i)}, 
\end{equation}
\noindent where $m_i$ and $a_i$ denote the numbers of missed and annotated teeth in image $i$, respectively. Missingness was $\leq 0.02$ in occlusal views and approximately 0.29 in other views, resulting an overall $\hat{p}_{missing}$ of 0.19 (95\% CI 0.17--0.21) among 1,577 teeth. Because non-annotated teeth are treated as background in the training process, this missingness may introduce label noise and degrade model performance. Digileap dataset correction was beyond the scope of this study; therefore, missing annotations were corrected only in the held-out test set.

\begin{figure*}[!t]
    \centering
    \includegraphics[width=\linewidth]{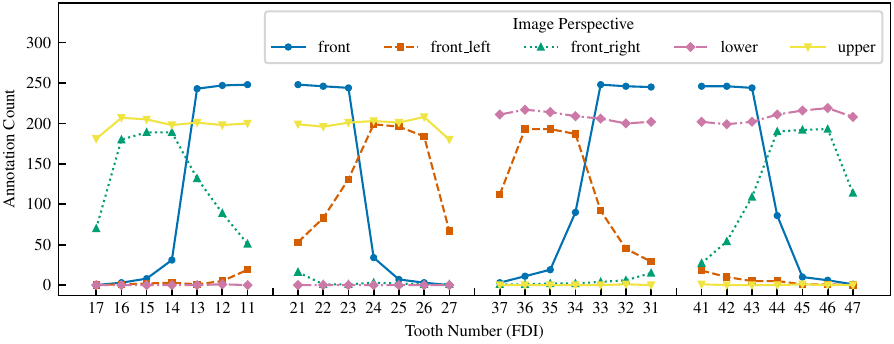}
    \caption{Annotated teeth count in each image perspective in training split.}
    \label{fig:digileap-view-and-teeth-dist}
\end{figure*}

To prevent data leakage, the dataset was randomly split at the participant level into training, validation, and test sets with image count ratios of 0.86:0.07:0.07, resulting in 1,094 training images and 89 images each for the validation and testing sets. All the images in the internal held-out test set were reviewed and missing annotations for tooth instances were added. The review was performed by the same annotator who did the annotation initially, and disagreements were resolved by S. Karki, a dentist with clinical expertise in oral health. The reviewer and the supervisor were unaware of the model predictions. As a result of this process, the annotation count of tooth instances in the test set increased from 1,143 to 1,365.
\begin{figure*}[t]
    \centering
    \includegraphics[scale = 1.0, width=1\linewidth]{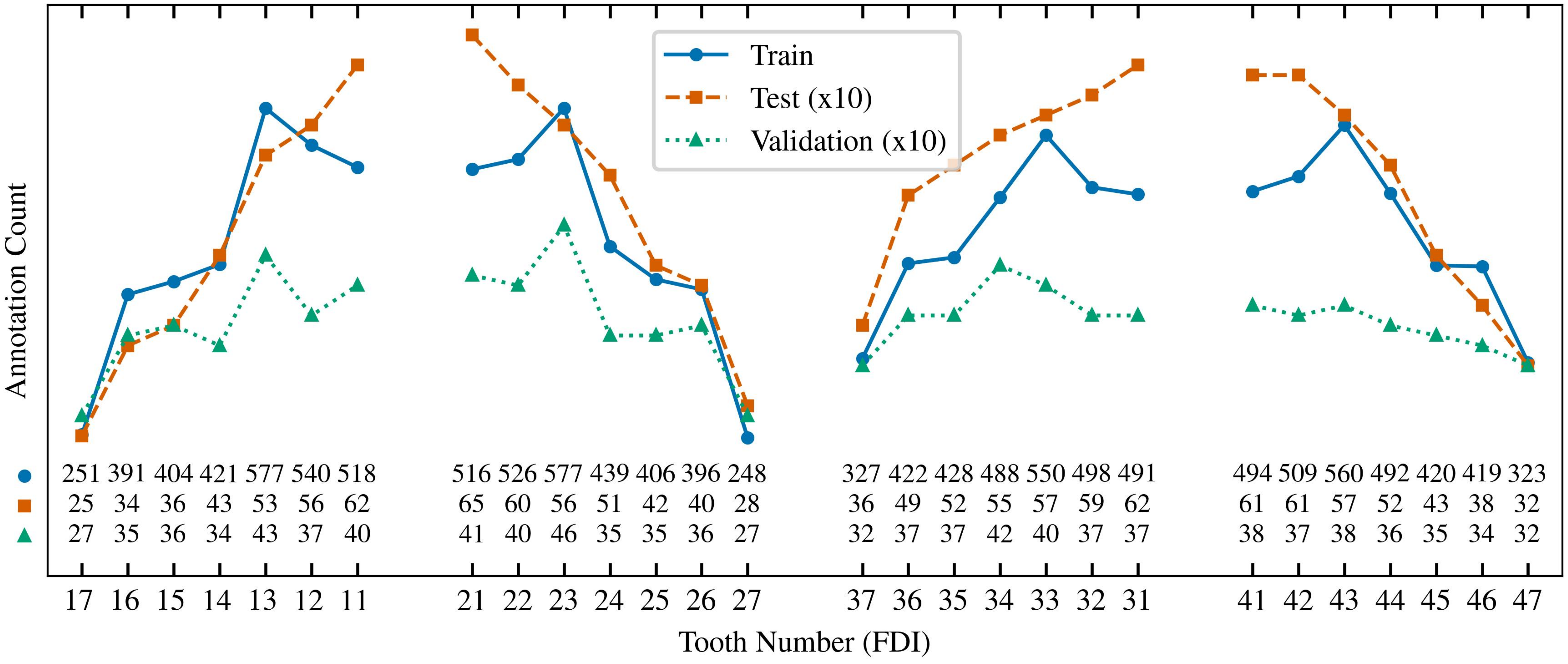}
    \caption{Distribution of tooth classes across training, validation, and test sets. The annotation counts for the validation and test sets were multiplied by 10 in the line plots to enhance the illustrative comparison. The actual values were printed in the three rows below the graphs. The test set is corrected missing annotations.}
    \label{fig:tooth-classes-distribution-across-splits}
\end{figure*}
The dataset partitions maintained overall consistent tooth class distributions slops, as shown in Figure \ref{fig:tooth-classes-distribution-across-splits}. Training and validation sets follow more similar patterns. However, there was a slightly lower incisor and canine counts ratio in the validation set compared to the training set, specifically in the lower jaw. The test set, includes a relatively greater number of teeth compared to the training and validation sets. This discrepancy arises primarily because missing annotations have been corrected in this subset, resulting in a deviation from the distribution pattern observed in the other sets.
\begin{table}[h] 
    \centering
    \caption{Distribution of Image View Completeness Among Participants.}
    \label{tab:participant_level_completeness}
    \begin{tabular*}{\linewidth}{@{\extracolsep{\fill}} ccc @{}}
        \toprule
        Image Count in the Set & Participants Count & Percentage \\
        \midrule
        1 & 19  & 5.83\%  \\
        2 & 37  & 11.35\% \\
        3 & 45  & 13.80\% \\
        4 & 81  & 24.85\% \\
        5 & 144 & 44.17\% \\
        \bottomrule
    \end{tabular*}
\end{table}
The DigiLeap dataset exhibited considerable variability in image dimensions, with widths ranging from 215 to 4,352 pixels and heights ranging from 127 to 3,591. This heterogeneity necessitated image resizing in the pipeline, which could introduce noise or information loss at the pixel level. In oral cavity images, the visible teeth in each view and the tooth surfaces visible vary significantly. Therefore, the inclusion of sufficient samples from anatomically and physically visible surfaces is important for the generalisability of the model and class-level performance. So, teeth that were not represented proportionally  (e.g. third molars and primary teeth) were excluded from the analysis. The shown in Figure \ref{fig:digileap-view-and-teeth-dist} in the training set anterior teeth dominated the frontal views, and premolars and molars were more present in the lateral and occlusal views.

The preprocessed dataset consisted of 1,272 images, consisting of 245 upper occlusal, 260 lower occlusal, 236 right lateral, 243 left lateral, and 288 frontal images of the oral cavity. 

\begin{algorithm}[!b]
\caption{Domain-Informed Masked Gray-World White Balance.}\label{alg:dental-gray-world}
\begin{algorithmic}
\STATE \textbf{Input:} RGB image $ \mathbf{I} \in [0,255]^{H \times W \times 3} $
\STATE \textbf{Output:} White-balanced RGB image $ \mathbf{I}^\prime $
\STATE 
\STATE {\textsc{DENTAL\_WHITE\_BALANCE}}$(\mathbf{I})$
\STATE \hspace{0.5cm}$ \mathbf{F} \gets \mathbf{I} / 255 $ 
\STATE \hspace{0.5cm}$ \mathbf{H} \gets \textsc{rgb2hsv}(\mathbf{F}) $
\STATE \hspace{0.5cm}\textit{// Blue gloves mask}
\STATE \hspace{0.5cm}$ \mathcal{M}_{\text{blue}} \gets (\mathbf{H}_h \in [0.43,0.70]) \land (\mathbf{H}_s \in [0.30,0.55]) $ 
\STATE \hspace{0.5cm}\textit{// Intra-oral voids mask}
\STATE \hspace{0.5cm}$ \mathcal{M}_{\text{void}} \gets (\mathbf{H}_h \notin (0.05,0.95)) \land (\mathbf{H}_s > 0.55) $ 
\STATE \hspace{0.5cm}$ \mathcal{M} \gets \mathcal{M}_{\text{blue}} \lor \mathcal{M}_{\text{void}} $
\STATE \hspace{0.5cm}$ V_{\text{avg}} \gets \textsc{mean}\big(\mathbf{H}_v \mid \neg \mathcal{M}_{\text{void}}\big) $
\STATE \hspace{0.5cm}$ R_{\text{avg}} \gets \textsc{mean}\big(\mathbf{F}_r \mid \neg \mathcal{M}\big) $
\STATE \hspace{0.5cm}$ G_{\text{avg}} \gets \textsc{mean}\big(\mathbf{F}_g \mid \neg \mathcal{M}\big) $
\STATE \hspace{0.5cm}$ B_{\text{avg}} \gets \textsc{mean}\big(\mathbf{F}_b \mid \neg \mathcal{M}\big) $
\STATE \hspace{0.5cm}\textit{// Masked global gray mean}
\STATE \hspace{0.5cm}$ \bar{G} \gets (R_{\text{avg}} + G_{\text{avg}} + B_{\text{avg}})/3 $ 
\STATE \hspace{0.5cm}\textit{// Brightness based gain calculation}
\STATE \hspace{0.5cm}$ f \gets 0.8 + 2\,(V_{\text{avg}} - 1)^2 $ 
\STATE \hspace{0.5cm}$ \mathbf{F}_r \gets \mathbf{F}_r \cdot (f+0.2) \cdot (\bar{G} / R_{\text{avg}}) $
\STATE \hspace{0.5cm}$ \mathbf{F}_g \gets \mathbf{F}_g \cdot f \cdot (\bar{G} / G_{\text{avg}}) $
\STATE \hspace{0.5cm}$ \mathbf{F}_b \gets \mathbf{F}_b \cdot f \cdot (\bar{G} / B_{\text{avg}}) $
\STATE \hspace{0.5cm}$ \mathbf{F} \gets \textsc{clip}(\mathbf{F}, 0, 1) $
\STATE \hspace{0.5cm}$ \mathbf{I}^\prime \gets \textsc{uint8}(\mathbf{F} \times 255) $
\STATE \hspace{0.5cm}\textbf{return} $ \mathbf{I}^\prime $
\end{algorithmic}
\end{algorithm}

\subsubsection{External Dataset}
Identifying an open dataset that matches DigiLeap in terms of imaging techniques, lighting, saturation, focus, and the severity of dental issues proved challenging. However, the "Teeth or Dental Image dataset" \cite{dixitchaudhary2024TeethDentalImage}, an open dataset provided by Bharati Vidyapeeth Dental College, India, was deemed appropriate for use as an external test set. Participants were aged 1–14 years. The image-capturing protocol used a mirror for occlusal views, and all images were captured using an iPhone 15$^\text{\textregistered}$ smartphone. Forty images initially met the selection criteria, but one was excluded during analysis because of incomplete annotation, leaving 39 images for external evaluation. Selection criteria was including only permanent dentition, equivalent severity of dental issues, and comparable artefacts and noise levels to those in DigiLeap. Dentistry students performed selection and annotation under supervision of dental experts S.K and E.V. Annotations followed the same protocol as Digileap and were checked to not to contain missing annotations. Each image in the resulting external dataset contained 20$\pm$1 annotated teeth.

\subsection{Pipeline and Model Architecture}

The preprocessing pipeline included a domain-informed masked gray-world white-balancing step to balance colours in each image.  Standard gray-world algorithms fail when large, uniform objects violate the achromatic scene assumption \cite{liImprovedGrayWorld}. To address this issue for oral cavity images, this study introduced a custom masking heuristic to excludes pixels corresponding to blue medical gloves and dark oral cavity voids prior to estimating the illuminant which is shown in Algorithm \ref{alg:dental-gray-world}. In the next step images were padded and resized to 1024×1024 pixels while maintaining their aspect ratio. Subsequently, the RGB channels were normalized using MS COCO dataset \cite{lin2014MicrosoftCOCOCommon} statistics to facilitate transfer learning with pre-trained backbone weights from a model that was trained on that dataset. During the training phase, augmentation applied by random brightness scaling, with a factor range of 0.7 to 1.1, and rotations within ±10 degrees were applied, each with a probability of 20\%.

This study employed Mask R-CNN \cite{he2020MaskRCNN}, a two-stage image segmentation and classification DL model. Apart from this model's success in delineating individual tooth boundaries in radiographic imagery \cite{wang2024CoMaskRCNNCollaborative}, Mask R-CNN allowed for the integrated implementation and evaluation of our domain-informed heuristic modifications. The specific implementation used in this study was a fork from the GitHub repository by W. Abdulla \cite{matterport_maskrcnn_2017}. This model can achieve high accuracy in object detection and image segmentation, and in addition to the bounding boxes and classes of objects, it provides polygon masks for the detected objects. Mask R-CNN is a hybrid model and it uses a hybrid loss function, defined as,

\begin{align}
\label{eq:loss}
    Loss &= \omega_{rc}L_{rpn\_cls} + \omega_{rb} L_{rpn\_bbox} + \nonumber \\
         &\quad \omega_{c} L_{cls} + \omega_{b} L_{bbox} + \omega_{m} L_{mask}.
\end{align}
The loss function is a weighted sum of components grouped by the three functional modules. For the region proposal network, the RPN classification loss ($L_{rpn\_cls}$) uses categorical cross-entropy to distinguish foreground from background, while the RPN bounding box loss ($L_{rpn\_bbox}$) uses Smooth L1 loss for regression to positive anchors. For detection and classification, the RoI classification loss ($L_{cls}$) is multiclass cross-entropy, and the RoI bounding box loss ($L_{bbox}$) uses Smooth L1 loss on positive RoIs. The mask loss ($L_{mask}$) is pixelwise binary cross-entropy for positive RoIs corresponding to the corresponding ground-truth class. Each component's contribution is controlled by weights in the model configuration. These weights were defined through hyperparameter tuning to balance detection, localisation and segmentation performance.
\begin{figure*}[t]
    \centering
    \includegraphics[width=\linewidth]{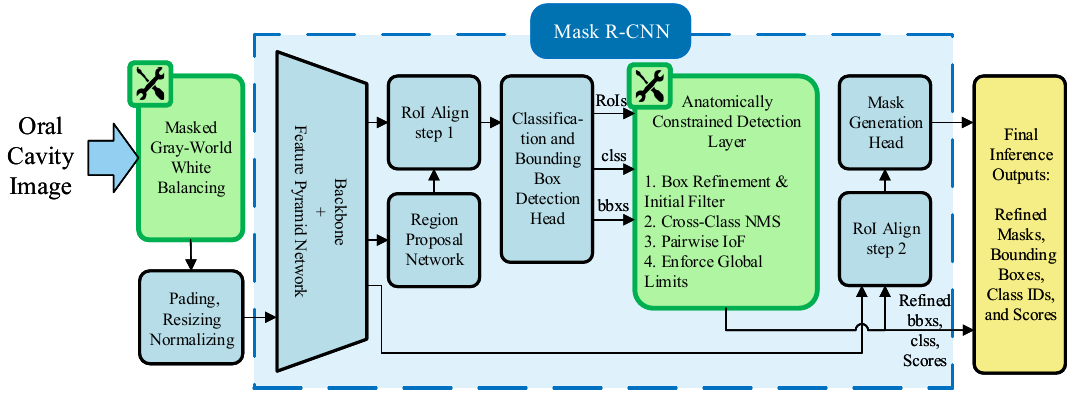} 
    \caption{Inference Pipeline that was developed for this study. Blocks in green are contributions from this research.}
    \label{fig:inference-pipeline}
\end{figure*}
\begin{algorithm}[!b]
\caption{Anatomically Constrained Detection Layer.}\label{alg:detection-layer}
\begin{algorithmic}
\STATE \textbf{Input:} ROIs $R$, Class Probabilities $P$, Box Deltas $D$; Thresholds $T_{conf}, T_{nms}^{class}, T_{nms}^{global}, T_{iof}$; Limits $K_{class}, K_{total}$
\STATE \textbf{Output:} Final detections $D_{final}$
\STATE 
\STATE {\textsc{DENTAL\_DETECTIONS}}$(R, P, D)$
\STATE \hspace{0.5cm}\textit{// Refine and clip to window}
\STATE \hspace{0.5cm}$ R \gets \textsc{apply\_deltas}(R, D) $
\STATE \hspace{0.5cm}$ S \gets \{ (r, p) \in R \times P \mid \text{argmax}(p) \neq$
\STATE \hspace{1.5cm}$\text{background} \land \max(p) \ge T_{conf} \} $
\STATE \hspace{0.5cm}$ S_{class} \gets \emptyset $
\STATE \hspace{0.5cm}\textbf{for each} unique class $c$ in $S$ \textbf{do}
\STATE \hspace{1.0cm}$ R_c \gets \{ r \in S \mid \textsc{class}(r) = c \} $
\STATE \hspace{1.0cm}\textit{// Strict class-wise Non-Maximum Suppression}
\STATE \hspace{1.0cm}$ R_c^\prime \gets \textsc{nms}(R_c, T_{nms}^{class}) $
\STATE \hspace{1.0cm}$ S_{class} \gets S_{class} \cup \textsc{topk}(R_c^\prime, K_{class}) $
\STATE \vspace{0.5pt}
\STATE \hspace{0.5cm}\textit{// Global cross-class Non-Maximum suppression}
\STATE \hspace{0.5cm}$ S_{global} \gets \textsc{nms}(S_{class}, T_{nms}^{global}) $

\STATE \hspace{0.5cm}\textit{// Intersection-over-Foreground filtering for nested detections suppression}
\STATE \hspace{0.5cm}\textbf{sort} $S_{global}$ by score descending
\STATE \hspace{0.5cm}\textbf{for each} $B_j \in S_{global}$ \textbf{do}
\STATE \hspace{1.0cm}\textbf{if} $ (\exists B_i \in S_{global}, i < j $ 
\STATE \hspace{1.5cm}such that $ \frac{\textsc{area}(B_i \cap B_j)}{\min(\textsc{area}(B_i), \textsc{area}(B_j))} > T_{iof}) $
\STATE \hspace{1.0cm}\textbf{then}
\STATE \hspace{1.5cm}$ S_{global} \gets S_{global} \setminus \{ B_j \} $
\STATE \vspace{0.5pt}
\STATE \hspace{0.5cm}$ D_{final} \gets \textsc{topk}(S_{global}, K_{total}) $
\STATE \hspace{0.5cm}\textbf{return} $ D_{final} $
\end{algorithmic}
\end{algorithm}

\subsubsection{Dental Detection Layer}
Although the loss function governs the training process, the model predictions can still be implausible based on domain knowledge. Therefore, the potential utility of the model in healthcare context requires that final predictions align with anatomical characteristics. To address this issue, we modified the detection layer of the Mask R-CNN implementation to incorporate prior anatomical constraints. The detection layer is a non-trainable layer that receives the ROIs, classification probabilities and bounding box regressions from the classification and bounding box heads, as shown in Figure \ref{fig:inference-pipeline}. As explained in Algorithm \ref{alg:detection-layer} this layer obtains a probability distribution for each candidate ROI from the classification branch of the head network. The class with the highest probability is assigned as the predicted label with its associated confidence score. Regions classified as background or with confidence below the predefined threshold were discarded. Because each tooth can appear at most once in an oral cavity image, detection is constrained to one region per class. To enforce this, NMS is applied within each class using a low threshold ( $IoU \geq 0.1$) to select the highest-scoring proposal. Because the model was trained on 28 tooth classes in this study, a maximum detection limit of 28 was automatically applied. However, to support training on a wider range of tooth classes, such as primary teeth, a second global constraint limits the total number of detections to the maximum number found in complete dentition. Therefore, in the second step, detections across all classes were merged and globally limited to 32. For this global stage, a higher NMS threshold ($IoU \geq 0.7$) is employed to accept more overlap than class-wise pruning.

In addition to the standard IoU-based suppression, an intersection over foreground (IoF) filtering step was introduced to improve the robustness of the model. It eliminates candidate boxes almost entirely contained within higher-scoring detections, which is anatomically implausible for tooth localisation because no valid tooth can be nested within another. This mechanism is particularly beneficial for patients wearing braces or other dental ornaments. It also reduces labelling workload and the number of training classes by eliminating the need to annotate decorative or metallic components. Overall, this two-stage suppression and constraint scheme acts as a domain-informed filtering step in the detection layer. The dental detection layer does not add parameters or affect the gradient flow because it is a non-differentiable component external to the computational graph. Thus, it indirectly influences training outcomes because the validation metrics that guide hyperparameter optimisation are computed after applying filtering.

\subsection{Training Strategy}
The model training involved optimising the weights based on the loss function. Nevertheless, training a hybrid model such as Mask R-CNN requires a substantial amount of data. This challenge is linked to the backbone block, which is a feature generator that comprises tens of millions of parameters. This study addressed the constraints of the limited DigiLeap dataset by employing transfer learning. For this purpose, the weights of the backbone feature generator were initialised from a pretrained ResNet101 network trained on the MS COCO dataset \cite{lin2014MicrosoftCOCOCommon}. All experimental procedures, hyperparameter tuning, and final training phases used transfer learning.

The Mask R-CNN model represents a framework with configuration parameters that can be refined according to the dataset characteristics and application domain properties. For instance, in this study, decisions regarding the dimensions and aspect ratios of the bounding boxes suggested by the region proposal network were made to guarantee the coverage of potential tooth instances within the dataset images. In this case, the minimum and maximum sizes of the ground-truth bounding boxes and the range of their aspect ratios served as the guides.

The hyperparameter optimisation phase systematically tuned the parameters, including the learning rate, weight decay, momentum, and loss function weights in Equation \ref{eq:loss}. A multistage strategy was employed to maintain computational feasibility. Initially, the backbone depth was evaluated using preliminary experiments. Hyperparameter tuning was then performed using the KerasTuner framework \cite{omalleytom2019KerasTuner}, employing Bayesian optimisation based on a Gaussian process. This adaptive approach balances exploration and exploitation, allowing quicker convergence to optimal configurations than grid or random searches, thereby minimising computational demands.

The final training was performed in two stages. The first stage restricted training to the network head weights for 10 epochs to stabilise the model. In the second stage, the training  was extended by 40 epochs which included the highest block of the backbone feature extractor. This fine-tuning phase used a reduced base learning rate combined with a learning rate annealing schedule to ensure precise weight adjustments.

\subsection{Evaluation Protocol}
The prediction pipeline was assessed as a system for segmenting tooth instances labelled with FDI tooth number. Each oral cavity photograph may contain a variable number of visible teeth. For the Digileap dataset, the participants were treated as independent sampling units in the participant-level analyses. However, participant identifiers were unavailable for the external dataset; therefore, images were treated as independent sampling units for the external validation.

The evaluation comprised confidence ranking and fixed operating point analyses. Predictions were exported at a minimum confidence threshold of 0.00 to retain all predicted instances. The complete ranked prediction set was used for the MS COCO evaluation protocol \cite{lin2014MicrosoftCOCOCommon}, whereas fixed operating point metrics were calculated using a confidence threshold selected exclusively from the validation set, as described below.

\subsubsection{Object Detection and Instance-Segmentation Metrics}
Confidence-ranked object detection and instance segmentation performances were evaluated based on the MS COCO evaluation protocol using the pycocotools \cite{wu2026PpwwyyxxCocoapi} implementation. Bounding box and mask average precision were reported as $\text{AP}_{50:95}$, averaged over IoU thresholds from 0.50 to 0.95 in increments of 0.05, together with $\text{AP}_{50}$ and $\text{AP}_{75}$. Individual tooth numbers were represented as object classes; therefore, class-wise $\text{AP}_{50}$ values for confidence-ranked predictions were also reported.

\subsubsection{Adapted instance Panoptic Quality}
End-to-end tooth localisation, numbering, and segmentation were evaluated using the tooth number-aware adapted instance panoptic quality with 
\begin{equation}
    \mathrm{PQ} = \frac{\sum_{(p,g) \in \mathrm{TP}} \operatorname{IoU}(p,g)}{\lvert \mathrm{TP} \rvert + 0.5 \lvert \mathrm{FP} \rvert + 0.5 \lvert \mathrm{FN} \rvert} = \mathrm{SQ} \times \mathrm{RQ},
    \label{eq:placeholder_label}
\end{equation}
\noindent Where
\begin{equation}
    \mathrm{SQ} = \frac{\sum_{(p,g) \in TP} \text{IoU}(p,g)}{|TP|}
    \label{eq:placeholder_label}
\end{equation}
\noindent represents segmentation quality among correctly matched instances, and
\begin{equation}
    \label{eq:placeholder_label}
    \mathrm{RQ} = \frac{\lvert TP \rvert}{\lvert TP \rvert + 0.5 \lvert FP \rvert + 0.5 \lvert FN \rvert}
\end{equation}
\noindent represents recognition quality and penalises missed and additional instances.

For the adapted PQ, predictions and ground truth instances were matched independently within each image using a one-to-one bipartite assignment. Eligible pairs had a mask $\mathrm{IoU>0.50}$. The assignment first maximised the number of matched pairs and then their total IoU, using SciPy’s linear assignment solver based on Crouse et al. \cite{crouse2016Implementing2DRectangular}. This adaptation was required because tooth-instance masks could overlap, unlike the non-overlapping panoptic partitions for which matching is automatically unique at $\mathrm{IoU>0.50}$. The PQ formulation and its decomposition into the SQ and RQ components followed the method described by Kirillov et al. \cite{kirillov2019PanopticSegmentation}.

Two complementary PQ scopes were calculated. The first was \textbf{class-aware adapted PQ}, in which matching was restricted to predictions and references with identical tooth numbers. A spatially correct tooth assigned an incorrect number therefore, contributed a false negative to its reference class and a false positive to its predicted class. The second was \textbf{class-agnostic tooth-instance PQ}, in which matching was based only on mask overlap and did not require tooth number agreement. This measured tooth localisation and segmentation independent of numbering. 

For the class-aware adapted PQ metric, the summed IoU, true positive(TP), false positive (FP), and false negative (FN) counts were pooled across the test set separately for each tooth number class. PQ, SQ, and RQ were calculated for each class and then macro-averaged, giving equal weight to each tooth class. Classes without references or predictions were excluded.

\subsubsection{Object Detection Metrics at Fixed Operating Point}
The operational precision, recall, and F1 were calculated after retaining predictions with confidence scores not less than the selected operating threshold. Predictions were processed in descending confidence order and matched to reference teeth using the class-aware Mask R-CNN matching procedure. A TP required an identical tooth number and mask $\mathrm{IoU>0.50}$. Unmatched predictions and references were counted as FPs and FNs, respectively.

Notably, although RQ has the algebraic form of F1 when both are derived from identical matching counts, the F1 and RQ values use different matching procedures and threshold inequalities. Therefore, they were not assumed to be identical.

\subsubsection{Matched-instance Dice and Tooth Numbering Metrics}
In addition to SQ, the segmentation quality was evaluated using the Dice similarity coefficient which follows Dice \cite{dice1945MeasuresAmountEcologic} and its application and interpretation in medical-image segmentation were informed by Taha and Hanbury \cite{taha2015MetricsEvaluating3D}. This metric is defined as
\begin{equation}
    \label{eq:placeholder_label}
    \text{DSC(P,G)} = \frac{ 2\lvert P \cap G\rvert}{\lvert P \rvert + \lvert G\rvert}=\frac{2IoU(P,G)}{1 + IoU(P,G)},
\end{equation}
\noindent where P and G denote the predicted and reference masks, respectively.

Predictions with confidence scores larger than or equal to the fixed operating threshold were matched to ground truth using greedy, class-agnostic, one-to-one matching at mask $\mathrm{IoU\geq0.50}$. Candidate pairs were prioritised first by greater IoU and then by prediction confidence. The Dice metric was calculated only for geometrically matched instances, irrespective of whether the predicted tooth number was correct. Therefore, the mask quality was measured after successful tooth localisation, and we called it matched-instance Dice. 

\subsubsection{Operating Point Selection}
A fixed confidence threshold was selected using the validation set. Confidence thresholds from 0.00 to 1.00 were evaluated in increments of 0.01 while keeping the matching IoU threshold fixed at 0.50. The selection criterion was the participant macro, class-agnostic tooth instance PQ. This procedure selected a confidence threshold of 0.27, which was fixed before evaluating the internal and external test sets. Because the selection criterion was class-agnostic, operating point selection prioritised tooth localisation and segmentation quality over tooth number correctness.

\subsubsection{Uncertainty Estimation}
To estimate uncertainty, 1000 bootstrap samples were generated from the participant clusters using a fixed random seed. During each iteration, participant identifiers were sampled with replacement. Metrics were recalculated for each resample, and two-sided 95\% percentile confidence intervals were derived from the bootstrap distributions. This resampling procedure maintained the dependence between multiple images associated with each participant.

\subsubsection{Internal, External, Ablation and Training Stability Analyses}
The final model was evaluated separately on the internal and external test sets using the confidence threshold selected during operating point selection. A descriptive ablation analysis using the internal test set assessed the contributions of masked gray-world white balancing and the anatomically constrained detection layer. All variants used the same inference procedure, confidence threshold, IoU criteria, and metric definitions as the complete pipeline. Therefore, the results were interpreted as analyses of components rather than independent estimates of generalisation.

Training stability was assessed by dividing the development data into ten folds and training ten models, each on a different nine-fold subset, such that every fold was omitted once. All models were trained for 50 epochs under identical settings and evaluated using the same reserved internal test set. The results were used to describe the sensitivity to the development data composition within the internal domain. Because the omitted fold was not used for evaluation, this was not a conventional cross-validation. Moreover, the overlapping training sets and shared test set provided correlated descriptive estimates of stability rather than independent estimates of generalisation.

\subsection{Experimental Setup}
Training and Inference were performed on a workstation equipped with an Nvidia$^{\text{\textregistered}}$ GeForce GTX 1080 GPU, an Intel$^{\text{\textregistered}}$ Core$^{\text{\texttrademark}}$ i7-8700 CPU, and 32GB of RAM. All experiments in this study were implemented using Python v3.10.12 \cite{PythonReleasePythona}, TensorFlow v2.14.0 \cite{tensorflow2015-whitepaper}, and accompanying Keras v2.14.0 \cite{chollet2015keras} frameworks. Because oral cavity images contain personal data, the DigiLeap dataset was accessed via an encrypted drive in accordance with data protection regulations.

\section{Results}
\label{results}
\begin{figure}[!t]
    \centering
    \includegraphics[width=1\linewidth]{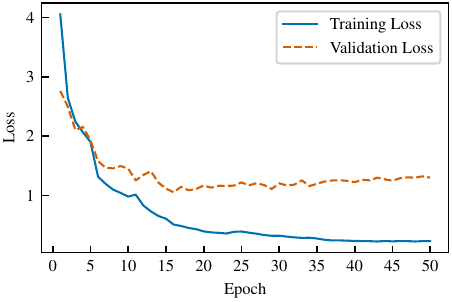}
    \caption{Training and validation loss graph.}
    \label{fig:train-val-loss-graph}
\end{figure}
This section presents the quantitative performance evaluation of our proposed pipeline. First, the hyperparameters selected during the tuning process are reported. Second, the model performance results are presented using two evaluation strategies: internal testing on held-out data and external validation on the independent dataset. Subsequently, the descriptive ablation study results show how the masked gray-world white balancing and dental detection modules affect the model performance. Finally, the training stability analysis results show the sensitivity of the pipeline to the internal domain data composition.
\begin{table}[htbp]
    \centering
    \caption{Hyperparameter search space and selected configurations which are marked in bracket and bold font. }
    \label{tab:hyperparameters}
    \begin{tabular*}{\columnwidth}{@{\extracolsep{\fill}} l l}
        \toprule
        \textbf{Hyperparameters} & \textbf{Search Space} \\ 
        \midrule
        Backbone & [\textbf{ResNet50}], ResNet101 \\ 
        Learning Rate & 0.0001, 0.0005, 0.001, [\textbf{0.006}], 0.01 \\ 
        Weight Decay & 0.001, [\textbf{0.0001}], 0.0005 \\ 
        Learning Momentum & 0.7, 0.75, [\textbf{0.8}], 0.9, 0.98 \\ 
        $\omega_c$ & 0.5, 1, 1.5, [\textbf{3}] \\ 
        $\omega_{rc}$ & 0.5, 1, 1.5, [\textbf{3}] \\ 
        $\omega_{rb}$ & 0.5, 1, 1.5, [\textbf{3}] \\ 
        $\omega_m$ & 0.5, [\textbf{1}], 1.5, 3 \\ 
        $\omega_b$ & 0.5, 1, 1.5, [\textbf{3}] \\ 
        \bottomrule
    \end{tabular*}
\end{table}

\begin{table*}[t]
    \centering
    \caption{Overall performance of the final framework on internal and external test sets. Values in brackets are 95\% confidence intervals.}
    \label{tab:overall_performance}
    \setlength{\tabcolsep}{3.2pt}
    \renewcommand{\arraystretch}{1.25}
    \resizebox{\textwidth}{!}{
    \begin{tabular*}{\textwidth}{@{\extracolsep{\fill}} l c ccc ccc ccc}
        \toprule
        \multirow{2}{*}{Evaluation}
        & \multicolumn{1}{c}{Bounding Box}
        & \multicolumn{3}{c}{Instance-Mask AP}
        & \multicolumn{3}{c}{Class-Aware Adapted PQ}
        & \multicolumn{3}{c}{Operational Classification} \\
        \cmidrule(lr){2-2}
        \cmidrule(lr){3-5}
        \cmidrule(lr){6-8}
        \cmidrule(lr){9-11}
        & $AP_{50:95}$
        & $AP_{50:95}$
        & $AP_{50}$
        & $AP_{75}$
        & PQ
        & SQ
        & RQ
        & Prec.
        & Recall
        & F1 \\
        \midrule

        Internal test set
        & \makecell{0.632\\{[0.609, 0.665]}}
        & \makecell{0.645\\{[0.618, 0.682]}}
        & \makecell{0.818\\{[0.784, 0.851]}}
        & \makecell{0.774\\{[0.742, 0.813]}}
        & \makecell{0.780\\{[0.755, 0.805]}}
        & \makecell{0.884\\{[0.874, 0.893]}}
        & \makecell{0.882\\{[0.854, 0.907]}}
        & 0.959
        & 0.821
        & 0.884 \\\\

        External test set
        & \makecell{0.708\\{[0.677, 0.747]}}
        & \makecell{0.744\\{[0.712, 0.780]}}
        & \makecell{0.901\\{[0.861, 0.940]}}
        & \makecell{0.893\\{[0.851, 0.932]}}
        & \makecell{0.832\\{[0.801, 0.857]}}
        & \makecell{0.896\\{[0.891, 0.901]}}
        & \makecell{0.928\\{[0.895, 0.957]}}
        & 0.943
        & 0.913
        & 0.928 \\
        \bottomrule
    \end{tabular*}
    }
    \vspace{2pt}
\end{table*}
\subsection{Hyperparameter Optimization Results}
Parameter search space exploration showed that the ResNet101 backbone did not provide a meaningful benefit to the model performance compared to ResNet50. Therefore, the smaller backbone was selected to decrease the computational cost. The search space and final optimal hyperparameters are presented in Table \ref{tab:hyperparameters}. The training and validation loss curves  for training the model using these parameters are shown in Figure \ref{fig:train-val-loss-graph}. The checkpoint at epoch 20 was selected as the final model for performance analyses, because the validation loss showed a slight positive slope after this training step, suggesting that overfitting had occurred.

\subsection{Evaluation results}
\subsubsection{Operating Point Selection on Validation dataset}
The validation performance remained unchanged for evaluation score thresholds from 0.00 to 0.27, with participant macro tooth-instance PQ equal to \(0.797\). This plateau occurred because the lowest retained prediction confidence was \(0.2764\); therefore, all 1,119 predictions were included by this threshold range. Performance began to decrease at thresholds above 0.27, as predictions were progressively excluded. Following the predefined rule of selecting the highest threshold to attain the maximum objective value, an evaluation score threshold of 0.27 was chosen.

\subsubsection{Internal Test Set Performance}
The held-out internal test set consisted of 89 images from the DigiLeap dataset. TLNM achieved a bounding box $\text{AP}_{50:95}$ of 0.632 and an instance mask $\text{AP}_{50:95}$ of 0.645, whereas the instance mask $\text{AP}_{50}$ and $\text{AP}_{75}$ were 0.818 and 0.774, respectively. At the confidence threshold of 0.27, the class-aware adapted PQ was 0.780 and the precision, recall, and F1 were 0.959, 0.821, and 0.884, respectively. The substantially higher precision than recall indicates that the model generated relatively few false-positive detections, while missed reference teeth were more common. The overall and class-wise evaluation results for the held-out test set are presented in Tables \ref{tab:overall_performance} and \ref{tab:classwise_results}.

\begin{figure}[t]
    \centering
    \includegraphics[width=\linewidth]{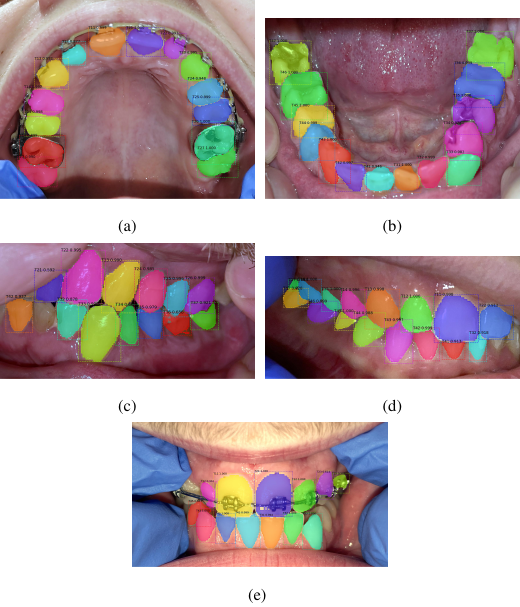}
    \caption{Sample predictions from the internal test set of the DigiLeap dataset. (a) Upper view, (b) Lower view, (c) Front-left view, (d) Front-right view, (e) Front view.}
    \label{fig:digileap-internal-testset-prediction-samples}
\end{figure}

TLNM showed strong operational performance across most tooth classes, with F1 $\geq 0.850$ in 21 of 28 classes; between-class variation was greatest for segmentation $\mathrm{AP}_{50}$, whereas class-aware PQ and F1 were more consistent. Tooth 23 achieved the highest values across all three metrics. Among molars, first molars consistently achieved higher class-aware PQ and F1 than the corresponding second molars in all four quadrants, whereas no consistent pattern was observed for segmentation $\mathrm{AP}_{50}$. Canines also showed consistently high performance across all three metrics. Performance was lower for mandibular incisors than for maxillary incisors, with teeth 31 and 41 among the weaker classes. Tooth 35 showed the lowest segmentation $\mathrm{AP}_{50}$ and class-aware PQ, while teeth 41 and 35 had the lowest F1 values. Figure \ref{fig:digileap-internal-testset-prediction-samples} presents TLNM detection, segmentation and numbering on samples from the held-out internal test set.

\subsubsection{External Test Set Performance}
The evaluation results for the external test set, presented in Table \ref{tab:overall_performance}, show that the TLNM scored higher in all metrics for this dataset, except for operational precision which was slightly lower. This result indicates that performance was maintained under the evaluated domain shift, although differences in dataset composition may also have contributed to this result. 

The class-wise evaluation results for the external test set are presented in Table \ref{tab:classwise_results}. On external validation, TLNM maintained strong performance across most tooth classes, with segmentation $\mathrm{AP}_{50}\geq0.850$ in 21 of 28 classes and operational F1 $\geq0.850$ in 27 of 28 classes. The performance was generally higher and more consistent for posterior teeth, whereas the anterior classes showed greater variability. Teeth 46, 34, 33, and 32 achieved perfect segmentation $\mathrm{AP}_{50}$. Teeth 46 and 34 also achieved F1 $=1.000$ and tooth 46 had the highest class-aware PQ. Tooth 42 had the lowest segmentation $\mathrm{AP}_{50}$ (0.779), while tooth 12 had the lowest class-aware PQ (0.715) and F1 (0.810). Tooth 33 remained among the stronger classes across both test sets, whereas tooth 23, which showed the highest internal performance, exhibited lower performance externally.
\begin{table*}[t]
    \centering
    \caption{
    Class-wise segmentation, class-aware panoptic quality, and operational detection performance on the internal and external test sets. Within each test set, highest and lowest values for metrics are bold and underlined, respectively.  Confidence intervals were calculated but omitted for visual clarity.
    }
    \label{tab:classwise_results}
    \setlength{\tabcolsep}{2.8pt}
    \renewcommand{\arraystretch}{1.25}

    \resizebox{\textwidth}{!}{
    \begin{tabular*}{\textwidth}{
        @{\extracolsep{\fill}} c l ccc ccc
    }
        \toprule
        \multirow{2}{*}{\textbf{Number}}
        & \multirow{2}{*}{\textbf{Anatomical sub-group}}
        & \multicolumn{3}{c}{\textbf{Internal test set}}
        & \multicolumn{3}{c}{\textbf{External test set}} \\
        \cmidrule(lr){3-5}
        \cmidrule(lr){6-8}

        &
        & Seg. $AP_{50}$
        & Class-Aware PQ
        & Operational F1
        & Seg. $AP_{50}$
        & Class-Aware PQ
        & Operational F1 \\
        \midrule

        \multicolumn{8}{l}{\hspace{3em}\textbf{Molars}} \\

        17 & \multirow{4}{*}{\textit{Maxillary (Upper)}}
        & 0.954 & 0.722 & 0.906
        & 0.941 & 0.871 & 0.973 \\

        16 &
        & 0.879 & 0.810 & 0.923
        & 0.901 & 0.850 & 0.947 \\

        26 &
        & 0.901 & 0.796 & 0.923
        & 0.934 & 0.851 & 0.950 \\

        27 &
        & 0.817 & 0.728 & 0.885
        & 0.891 & 0.827 & 0.944 \\

        \addlinespace

        37 & \multirow{4}{*}{\textit{Mandibular (Lower)}}
        & 0.810 & 0.744 & 0.857
        & 0.974 & 0.865 & 0.973 \\

        36 &
        & 0.771 & 0.764 & 0.864
        & 0.997 & 0.893 & 0.974 \\

        46 &
        & 0.782 & 0.774 & 0.870
        & \textbf{1.000} & \textbf{0.913} & \textbf{1.000} \\

        47 &
        & 0.752 & 0.749 & 0.842
        & 0.941 & 0.875 & 0.974 \\

        \addlinespace

        \multicolumn{8}{l}{\hspace{3em}\textbf{Premolars}} \\

        15 & \multirow{4}{*}{\textit{Maxillary (Upper)}}
        & 0.802 & 0.800 & 0.892
        & 0.982 & 0.875 & 0.974 \\

        14 &
        & 0.791 & 0.785 & 0.872
        & 0.901 & 0.853 & 0.947 \\

        24 &
        & 0.737 & 0.749 & 0.826
        & 0.851 & 0.787 & 0.872 \\

        25 &
        & 0.820 & 0.771 & 0.875
        & 0.873 & 0.820 & 0.919 \\

        \addlinespace

        35 & \multirow{4}{*}{\textit{Mandibular (Lower)}}
        & \underline{0.672} & \underline{0.702} & 0.795
        & 0.941 & 0.888 & 0.973 \\

        34 &
        & 0.868 & 0.830 & 0.923
        & \textbf{1.000} & 0.910 & \textbf{1.000} \\

        44 &
        & 0.762 & 0.774 & 0.870
        & 0.871 & 0.817 & 0.895 \\

        45 &
        & 0.718 & 0.711 & 0.805
        & 0.934 & 0.820 & 0.900 \\

        \addlinespace

        \multicolumn{8}{l}{\hspace{3em}\textbf{Anterior Teeth}} \\

        13 & \multirow{2}{*}{\textit{Maxillary Canines}}
        & 0.919 & 0.846 & 0.942
        & 0.795 & 0.757 & 0.872 \\

        23 &
        & \textbf{1.000} & \textbf{0.896} & \textbf{1.000}
        & 0.842 & 0.795 & 0.889 \\

        \addlinespace

        33 & \multirow{2}{*}{\textit{Mandibular Canines}}
        & 0.911 & 0.850 & 0.954
        & \textbf{1.000} & 0.863 & 0.974 \\

        43 &
        & 0.850 & 0.808 & 0.916
        & 0.932 & 0.830 & 0.923 \\

        \addlinespace

        12 & \multirow{4}{*}{\textit{Maxillary Incisors}}
        & 0.909 & 0.856 & 0.944
        & 0.809 & \underline{0.715} & \underline{0.810} \\

        11 &
        & 0.851 & 0.826 & 0.914
        & 0.851 & 0.836 & 0.919 \\

        21 &
        & 0.780 & 0.771 & 0.850
        & 0.851 & 0.833 & 0.919 \\

        22 &
        & 0.851 & 0.822 & 0.911
        & 0.802 & 0.797 & 0.889 \\

        \addlinespace

        32 & \multirow{4}{*}{\textit{Mandibular Incisors}}
        & 0.828 & 0.797 & 0.899
        & \textbf{1.000} & 0.836 & 0.974 \\

        31 &
        & 0.692 & 0.723 & 0.811
        & 0.808 & 0.767 & 0.872 \\

        42 &
        & 0.761 & 0.739 & 0.847
        & \underline{0.779} & 0.761 & 0.857 \\

        41 &
        & 0.708 & 0.704 & \underline{0.793}
        & 0.832 & 0.781 & 0.865 \\

        \bottomrule
    \end{tabular*}
    }

    \vspace{2pt}
\end{table*}
\subsubsection{Ablation Study}
The descriptive ablation analysis showed complementary effects of the two proposed modules. Masked gray-world white balancing produced modest improvements over the vanilla configuration across all evaluated metrics, with the largest relative changes observed in confidence-ranked segmentation and recall. In contrast, the anatomically constrained detection layer increased precision and improved both tooth-instance and class-aware PQ and operational F1, while reducing recall and segmentation $\mathrm{AP}_{50:95}$.

Combining both modules retained the precision and PQ gains observed with the anatomically constrained detection layer while partially recovering the reductions in recall and segmentation $\mathrm{AP}{50:95}$. The final framework achieved the highest tooth-instance PQ, class-aware PQ, operational precision, and operational F1, although its segmentation $\mathrm{AP}{50:95}$ remained close to the vanilla baseline. The matched-instance Dice remained unchanged across the configurations. Overall, the ablation results showed that the principal performance differences between configurations occurred in detection, recognition, and operating-point behaviour rather than in the segmentation quality of successfully matched teeth.

\begin{table*}[!h]
    \centering
    \caption{Descriptive ablation analysis of masked gray-world white balancing (WB) and the anatomically constrained detection layer on the internal test set. Bold font indicates best performance.}
    \label{tab:ablation_analysis}
    \begin{tabular*}{\textwidth}{@{\extracolsep{\fill}} l cc c cc ccc c}
        \toprule
        \multirow{2}{*}{Configuration}
        & \multicolumn{2}{c}{Pipeline Modules}
        & \multicolumn{1}{c}{Ranked Segmentation}
        & \multicolumn{2}{c}{Adapted PQ}
        & \multicolumn{3}{c}{Operational Classification}
        & \multicolumn{1}{c}{Mask Quality} \\
        \cmidrule(lr){2-3}
        \cmidrule(lr){4-4}
        \cmidrule(lr){5-6}
        \cmidrule(lr){7-9}
        \cmidrule(lr){10-10}
        & WB
        & Detection Layer
        & Mask $AP_{50:95}$
        & Tooth PQ
        & Class PQ
        & Prec.
        & Recall
        & F1
        & matched-instance Dice \\
        \midrule
        Vanilla Mask R-CNN
        & Off & Original
        & 0.647
        & 0.777
        & 0.751
        & 0.874
        & 0.830
        & 0.852
        & 0.937 \\

        Masked White balancing only
        & On & Original
        & \textbf{0.656}
        & 0.785
        & 0.763
        & 0.892
        & \textbf{0.840}
        & 0.865
        & \textbf{0.938} \\

        Custom Detection layer only
        & Off & Custom
        & 0.637
        & 0.807
        & 0.771
        & 0.951
        & 0.811
        & 0.875
        & 0.937 \\

        Final framework
        & On & Custom
        & 0.645
        & \textbf{0.809}
        & \textbf{0.780}
        & \textbf{0.959}
        & 0.821
        & \textbf{0.884}
        & \textbf{0.938} \\
        \bottomrule
    \end{tabular*}

    \vspace{2pt}
\end{table*}

\subsubsection{Training Stability Analysis Results}
The fold-based training stability results are presented in Table \ref{tab:training_stability}. Despite the variation of 10\% of data points in the training set, performance on the common held-out internal test set varied only marginally across the ten models. The lower adapted panoptic segmentation quality indicates that the model learned segmentation more effectively than the other tasks. Overall, these findings show that the training pipeline remained stable with the data-partitioning scheme used in this study.

\begin{table*}[!h]
    \centering
    \caption{
        Fold-based training stability of the final framework. Values are reported as the mean $\pm$ standard deviation across
        the ten trained models.}
    \label{tab:training_stability}
    \setlength{\tabcolsep}{3.2pt}
    \renewcommand{\arraystretch}{1.25}

    \resizebox{\textwidth}{!}{
    \begin{tabular*}{\textwidth}{
        @{\extracolsep{\fill}} l c ccc ccc ccc
    }
        \toprule
        \multirow{2}{*}{Evaluation}
        & \multicolumn{1}{c}{Bounding Box}
        & \multicolumn{3}{c}{Instance-Mask AP}
        & \multicolumn{3}{c}{Class-Aware Adapted PQ}
        & \multicolumn{3}{c}{Operational Detection} \\
        \cmidrule(lr){2-2}
        \cmidrule(lr){3-5}
        \cmidrule(lr){6-8}
        \cmidrule(lr){9-11}

        & $AP_{50:95}$
        & $AP_{50:95}$
        & $AP_{50}$
        & $AP_{75}$
        & PQ
        & SQ
        & RQ
        & Prec.
        & Recall
        & F1 \\
        \midrule

        Training stability
        & \makecell{$0.626$\\ $\pm 0.007$}
        & \makecell{$0.636$\\ $\pm 0.008$}
        & \makecell{$0.798$\\ $\pm 0.009$}
        & \makecell{$0.765$\\ $\pm 0.006$}
        & \makecell{$0.776$\\ $\pm 0.004$}
        & \makecell{$0.887$\\ $\pm 0.002$}
        & \makecell{$0.875$\\ $\pm 0.004$}
        & \makecell{$0.966$\\ $\pm 0.008$}
        & \makecell{$0.796$\\ $\pm 0.008$}
        & \makecell{$0.873$\\ $\pm 0.004$} \\

        \bottomrule
    \end{tabular*}
    }

    \vspace{2pt}
\end{table*}

\section{Discussion}
In this study, we aimed to develop a deep learning framework for detecting, segmenting, and anatomically numbering teeth in smartphone images captured in uncontrolled environments. The modified domain informed Mask R-CNN pipeline demonstrated effective performance on both internal and external test sets. The masked gray-world white balancing preprocessing step improved ranked segmentation and recall; whereas the anatomically informed detection layer enhanced precision and end-to-end class aware performance. The higher external scores should be interpreted cautiously because the test sets differed in composition

\subsection{Methodological and Architectural Contributions}
Transitioning from clinical-grade imaging methods to PGHD, such as smartphone-captured oral cavity images, introduces engineering challenges due to non-standard and highly variable data quality. To address these challenges, we moved beyond the "black-box" implementation of deep learning models by embedding domain knowledge into the pipeline.

\subsubsection{Efficacy of Masked Gray-World White Balancing}
The ablation findings suggest that this preprocessing step provided a modest but broad benefit rather than driving a large change in any single aspect of performance. When applied alone, white balancing improved all evaluated metrics relative to the vanilla pipeline, with the most apparent gains in confidence-ranked segmentation, panoptic quality and recall. These findings are consistent with white balancing improving the robustness of learned visual representations to variations in colour and illumination. However, the ablation analysis did not identify the specific image characteristics responsible for the observed gains.

\subsubsection{Clinical Utility of the Anatomically Constrained Detection Layer}
The anatomically constrained detection layer produced a distinct performance profile, with precision increasing from 0.874 to 0.951 when applied alone, alongside improvements in tooth-instance PQ, class-aware PQ, and operational F1. These gains occurred despite modest reductions in recall and segmentation $\mathrm{AP}_{50:95}$. The unchanged matched-instance Dice suggests that the benefit resulted primarily from more selective instance detection rather than improved mask delineation. The anatomical constraints removed substantially more false-positive than true-positive predictions, improving prediction validity and tooth recognition at the selected operating point.

When combined with masked white balancing, the final framework retained the precision and PQ gains of the anatomical layer while recovering part of the loss in recall and ranked segmentation performance, yielding the highest operational F1. This supports the complementary roles of the two modules, with masked gray-world white balancing mainly improving visual robustness and sensitivity, and the anatomical layer improving prediction validity and precision. Because the ablation analysis was descriptive and limited to the internal test set, these findings should be interpreted as support for the design choices rather than independent evidence of generalisable benefit.

\subsection{Robustness, Domain Shifts, and Testing Paradigms}
\subsubsection{Analysis of Internal Training Stability}
The deployment of machine learning in healthcare requires reproducible training and limited sensitivity to changes in training data composition. The observed small between-model variations indicate that the training pipeline was stable under the evaluated changes in the development data composition. However, this does not imply that training is stable in other domains.

\subsubsection{Model Performance under External Domain Shifts}
Deep learning models in medical imaging may experience reduced performance under domain shifts caused by differences in image-capturing hardware, population characteristics, and acquisition protocols. In this study, however, performance was higher for several metrics on the independent external test set. The higher performance values observed for the external domain can be influenced by differences in image view composition, tooth visibility, average image quality, and class distribution. However, these findings indicate that the model retained effective performance under the evaluated domain shift. Two characteristics of the pipeline may have contributed to this result.

\begin{itemize}
    \item \textbf{Data Diversity:} The internal development dataset included substantial real-world variation in lighting, focus, smartphone sensors, orientation, and image quality. Training with this heterogeneous PGHD may have reduced the dependence on narrowly controlled acquisition conditions and encouraged the learning of features that generalised across image sources.
    \item \textbf{Preprocessing Strategy:} The masked gray-world white-balancing procedure was designed to reduce colour variation caused by differences in illumination, smartphone sensors and image processing algorithms. The ablation analysis showed modest improvements in the ranked segmentation performance and recall when white balancing was applied. Therefore, this preprocessing step may have contributed to greater colour consistency across datasets, although its specific effect on external test set performance was not evaluated independently.
\end{itemize}

\subsection{Granular Anatomical Performance Patterns}
\subsubsection{View-Dependent Performance in Posterior Teeth}
Analysis of the class-wise findings indicates that the enhanced external performance was predominantly concentrated in the posterior teeth, rather than being uniformly distributed across the entire dentition. The mean operational F1 increased from 0.884 to 0.967 for molars and from 0.857 to 0.935 for premolars, with corresponding improvements in segmentation and class-aware PQ. In contrast, anterior mean F1 remained unchanged between the internal and external datasets with 0.898 versus 0.897 mean values respectively.

This pattern should not be interpreted as evidence that posterior anatomy is easier for the model. The external dataset contained only upper and lower views, whereas the internal dataset included frontal and lateral views in addition to them. This data composition resulted in less variability in the visible tooth surfaces in the external dataset for the same anatomical classes. Therefore, the strong external posterior tooth group performance may reflect a more favourable or consistent visual presentation of these teeth in occlusal views, in addition to morphological distinctiveness. Consequently, tooth type and image-capturing view cannot be separated as explanations from the present evaluation.

\begin{figure}[t]
    \centering
    \includegraphics[width=\linewidth]{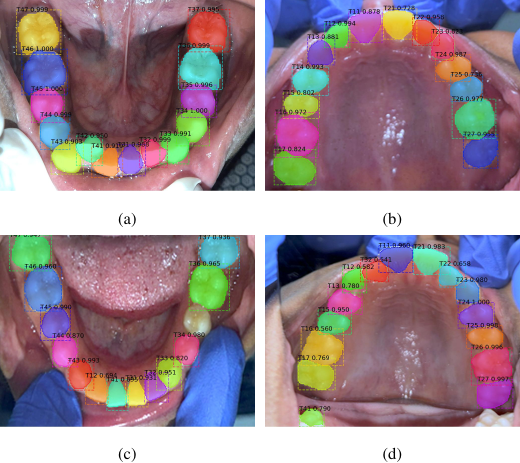}
    \caption{Sample predictions from the external test set. (a) Lower view, (b) Upper view, (c) Lower view, (d) Upper view.}
    \label{fig:digileap-externaltestset-prediction-samples}
\end{figure}

\subsubsection{Anterior Performance and Class-Specific Variability}
The anterior teeth showed a different pattern. Their average performance changed relatively little between the datasets, whereas the class-level variability was higher in both evaluations. Additionally, in the internal test set, the anterior teeth achieved a higher average performance than the posterior teeth, except for the mandibular classes. In the external test set this pattern was not visible, indicating that reduced performance was not consistently associated with either the maxillary or mandibular arch. These observations indicate that the viewing angle and visible tooth surfaces affect the success of the model in detection, segmentation, and numbering tasks.

\subsection{Limitations}
Despite these promising results, this study had four limitations. First, although the DigiLeap dataset comprises 1,272 images, it is relatively small compared with standard computer vision benchmarks. Second, TLNM is predominantly trained on permanent dentition, which may limit its applicability to mixed dentition in children. Third, the age distribution of the training dataset was skewed toward adolescents (comprising 229 secondary school students compared to only 97 adults) which may reduce the model's exposure to age-related dental conditions, extensive restorative work, or morphological changes that are more frequently encountered in older populations. Despite the independent review and correction of the internal test reference annotations before the final evaluation, missing tooth annotations remained a potential source of label noise in the development data.

\subsection{Operational Implications}
The model developed in this study functions as an automated indexing tool. This model can generate a structural dental map with automated localisation and numbering for each tooth which additionally provides segmentation masks. This capability is critical as a building block for remote dental screening and diagnostic system development. By transforming raw smartphone images into structured anatomical data, potential downstream diagnostic algorithms can focus exclusively on specific regions of interest (single teeth in oral cavity images).

This approach streamlines data collection and annotation tasks in teledentistry research. With the use of automated indexing, subsequent research can avoid the lengthy process of manually mapping teeth. Consequently, future studies targeting pathology detection, such as caries or plaque diagnosis, can benefit from this model. To support reproducibility and widespread adoption, the model was released as an open-source, containerised tool with a FastAPI interface, available on GitHub and Docker Hub. 

\section{Conclusion}
In this study, we developed TLNM, a deep learning pipeline for tooth localisation, instance segmentation, and anatomical numbering in highly variable dental PGHD. The integration of a masked gray-world white-balancing preprocessing method and an anatomically constrained detection layer reduced colour variability and suppressed anatomically implausible predictions. The ablation analysis showed that combined effect of both modules was increased operational precision and improved class-aware PQ and operational F1. Although these gains were accompanied by modest reductions in recall and ranked segmentation performance.

A comprehensive evaluation demonstrated high training stability in the internal dataset domain. On the held-out internal test set, TLNM achieved an instance-mask $\text{AP}_{50}$ of 0.818, class-aware PQ of 0.780, and operational F1 of 0.884. External evaluation further demonstrated that performance was maintained under differences in sensors, population, and image acquisition protocols, achieving an instance mask $\text{AP}_{50}$ of 0.901, a class-aware PQ of 0.832, and an operational F1 of 0.928. Although differences in dataset composition may have contributed to the higher external results, these findings support the generalisability of the framework under the evaluated domain shift.

To support reproducibility and the development of automated downstream diagnostic systems, the TLNM inference pipeline was released as an open-source containerised API. Overall, this framework demonstrates the feasibility of extracting tooth level anatomical information from images captured using consumer grade smartphones. Therefore, this approach provides a scalable and potentially cost-effective foundation for future remote oral health screening and teledentistry applications.

\section*{Code Availability}

To support open research, the TLNM API implementation is publicly available under the MIT licence. The source code is hosted in the GitHub repository \cite{nedaei2026H4ppy0wlTeeth_localization_and_numbering}, and a prebuilt Docker image of the tool is available via Docker Hub \cite{H4ppy0vvlTeeth_localization_and_numberingDocker}. Comprehensive documentation, including instructions for container deployment and API usage, is provided in the repository.

\section*{Acknowledgment}
The authors acknowledge the financial support of the European Regional Development Fund (ERDF) for funding the DigiLeap of Oral Health project (grant number EURA 2014/11292/09 02 01 01/2021/PPL), which facilitated the collection of the primary dataset utilized in this study. This research was also supported by the Minerva Foundation, the Finnish Dental Society Apollonia, and The Finnish Medical Foundation. We extend our gratitude to Katri Kukkola for her vital role in project management. We also thank the four dentists and two senior dental students who did the clinical annotations for the internal dataset, as well as the dentistry students who annotated the external validation dataset.

\section*{Author Contributions}
Conceptualisation: S.K, J.S, A.N; Methodology: A.N, H.T, J.S; Algorithm/Model development: A.N; Validation: A.N, S.K, E.V; Formal analysis: A.N; Data curation: A.N; Writing - original draft: A.N; Writing - review \& editing: H.T, J.S; Supervision: J.S, H.T; Funding acquisition: J.S, S.K.

\section*{Conflict of Interest}
The authors declare no conflict of interest.

\section*{Ethics \& Consent}
The data used in this study were initially collected in the DigiLeap of Oral Health project. Before the original data collection, approval for the study protocol was obtained from the ethical committee of the Northern Ostrobothnia Hospital District (EETTMK 62/2021), and the Finnish Medicines Agency (FIMEA) granted a Medical Device Permit (2022/007715). Additionally, permission for the original research was secured from public healthcare services in Kuusamo, Ylivieska, Oulu, and Liminka. The initial study adhered to the guidelines of the World Medical Association Declaration of Helsinki. Informed consent was obtained from parents or legal guardians of participants younger than 15 years. Participants older than 15 years provided informed consent themselves. Participation was entirely voluntary, and all participants were informed that they could withdraw from the study at any time without any adverse consequences. In this study, handling and access to the dataset were conducted in accordance with the permissions provided by the Finnish Social and Health Data Permit Authority, Findata, under permit number THL/6268/14.02.00/2021.

\section*{Data Availability}
The data used in this study comprised two datasets. The internal development dataset (DigiLeap) contains sensitive patient-generated health data and is subject to strict data protection regulations. Access to this dataset is restricted and governed by the permissions granted by the Finnish Social and Health Data Permit Authority (Findata). The external dataset used for validation is publicly available via the Mendeley Data repository under the title "Teeth or Dental Image dataset" \cite{dixitchaudhary2024TeethDentalImage}.

\section*{References}
\bibliographystyle{IEEEtran}
\bibliography{references}

\end{document}